\documentclass[10pt,twocolumn,letterpaper]{article}

\usepackage[pagenumbers]{cvpr} 

\usepackage[dvipsnames,table]{xcolor}
\usepackage{multirow}
\usepackage{makecell}

\renewcommand{\paragraph}[1]{\vspace{1.25mm}\noindent\textbf{#1}}

\definecolor{cvprblue}{rgb}{0.21,0.49,0.74}
\usepackage[pagebackref,breaklinks,colorlinks,citecolor=cvprblue]{hyperref}
\usepackage[accsupp]{axessibility}

\def\paperID{5435} 
\def\confName{CVPR}
\def\confYear{2024}

\title{RepViT: Revisiting Mobile CNN From ViT Perspective}

\author{Ao Wang$^{1,2}$ \quad Hui Chen$^{1,2}$\thanks{Corresponding author.} \quad Zijia Lin$^{1}$ \quad Jungong Han$^{3}$ \quad Guiguang Ding$^{1,2}$\footnotemark[1] \\
		$^1$Tsinghua University \quad $^2$BNRist \quad $^3$The University of Sheffield \\
		{\tt\small wa22@mails.tsinghua.edu.cn \quad huichen@mail.tsinghua.edu.cn \quad linzijia07@tsinghua.org.cn} \\ 
		{\tt\small jungonghan77@gmail.com \quad dinggg@tsinghua.edu.cn }}

\begin{document}
\maketitle

\begin{abstract}
Recently, lightweight Vision Transformers (ViTs) demonstrate superior performance and lower latency, compared with lightweight Convolutional Neural Networks (CNNs), on resource-constrained mobile devices. Researchers have discovered many structural connections between lightweight ViTs and lightweight CNNs. However, the notable architectural disparities in the block structure, macro, and micro designs between them have not been adequately examined. In this study, we revisit the efficient design of lightweight CNNs from ViT perspective and emphasize their promising prospect for mobile devices. Specifically, we incrementally enhance the mobile-friendliness of a standard lightweight CNN, \ie, MobileNetV3, by integrating the efficient architectural designs of lightweight ViTs. This ends up with a new family of pure lightweight CNNs, namely RepViT. Extensive experiments show that RepViT outperforms existing state-of-the-art lightweight ViTs and exhibits favorable latency in various vision tasks. Notably, on ImageNet, RepViT achieves over 80\% top-1 accuracy with 1.0 ms latency on an iPhone 12, which is the first time for a lightweight model, to the best of our knowledge. Besides, when RepViT meets SAM, our RepViT-SAM can achieve nearly 10$\times$ faster inference than the advanced MobileSAM. Codes and models are available at \url{https://github.com/THU-MIG/RepViT}.
\end{abstract}

\section{Introduction}
\label{sec:intro}

In the field of computer vision, designing lightweight models has been a major focus for achieving superior model performance with reduced computational costs. This is particularly important for resource-constrained mobile devices to enable the deployment of visual models at the edge. Over the past decade, researchers have primarily focused on lightweight convolutional neural networks (CNNs) and have made significant progress. Many efficient design principles have been proposed, including separable convolutions~\cite{howard2017mobilenets}, inverted residual bottlenecks~\cite{sandler2018mobilenetv2}, channel shuffling~\cite{zhang2018shufflenet,ma2018shufflenet}, and structural re-parameterization~\cite{ding2021repvgg,ding2019acnet}, \etc. These design principles have led to the development of representative models like MobileNets~\cite{howard2017mobilenets,sandler2018mobilenetv2,howard2019searching}, ShuffleNets~\cite{ma2018shufflenet,zhang2018shufflenet}, and RepVGG~\cite{ding2021repvgg}.

\begin{figure}[t]
\centering
    \includegraphics[width=1.0\linewidth]{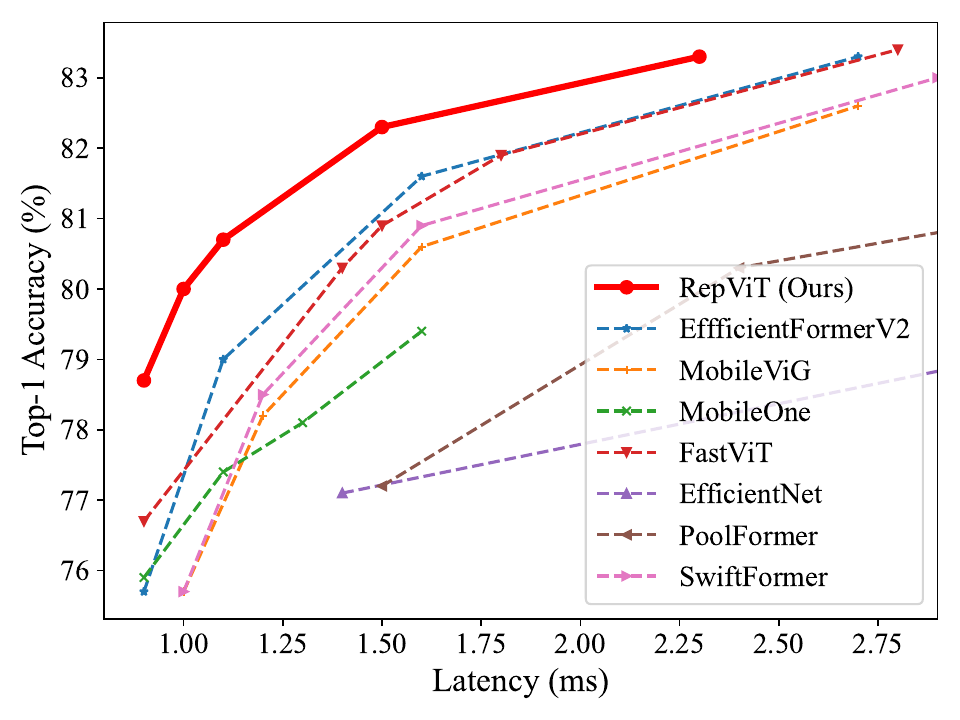}
    \caption{Comparison of latency and accuracy between RepViT (Ours) and others. The top-1 accuracy is tested on ImageNet-1K and the latency is measured by iPhone 12 with iOS 16. RepViT achieves the best trade-off between performance and latency.}
    \label{fig:intro}
    \vspace{-16pt}
\end{figure}

In recent years, Vision Transformers (ViTs)~\cite{dosovitskiy2020image} have emerged as a promising alternative to CNNs for learning visual representations. They have demonstrated superior performance compared to CNNs on a variety of vision tasks, like image classification~\cite{liu2021swin,wang2021pyramid}, semantic segmentation~\cite{xie2021segformer,cheng2022masked} and object detection~\cite{carion2020end,li2112improved}. However, the trend of increasing the number of parameters in ViTs to improve performance results in large model sizes and high latency~\cite{liu2022swin,dehghani2023scaling}, making them unsuitable for resource-constrained mobile devices~\cite{mehta2021mobilevit,li2022efficientformer}. Although it is possible to directly reduce the model size of ViT models to match the constraints of mobile devices, their performance often becomes inferior to that of lightweight CNNs~\cite{chen2022mobile}. Therefore, researchers have embarked on exploring the lightweight design of ViTs, aiming to achieve performance surpassing that of lightweight CNNs.

Many efficient design principles have been proposed to enhance the computational efficiency of ViTs for mobile devices~\cite{li2022rethinking,mehta2021mobilevit,chen2022mobile,pan2022edgevits}. Some approaches propose innovative architectures that combine convolutional layers with ViTs, resulting in hybrid networks~\cite{chen2022mobile,mehta2021mobilevit}. Additionally, novel self-attention operations with linear complexity~\cite{mehta2022separable} and dimension-consistent design principles~\cite{li2022efficientformer,li2022rethinking} are introduced to improve the efficiency. These studies demonstrate that lightweight ViTs~\cite{mehta2022separable,li2022rethinking,pan2022edgevits} can achieve lower latency on mobile devices while outperforming lightweight CNNs~\cite{vasu2023mobileone,sandler2018mobilenetv2,howard2019searching}, as shown in \Cref{fig:intro}.


Despite the success of lightweight ViTs, they continue to face practical challenges due to inadequate hardware and computational library support \cite{vasu2023mobileone}. Additionally, ViTs are susceptible to inputs with high resolution, resulting in high latency~\cite{bakhtiarnia2022efficient}. In contrast, CNNs utilize highly optimized convolution operations with linear complexity relative to the input, making them advantageous for deployment on edge devices \cite{zhang2022parc,sateesan2021survey}. Therefore, designing high-performance lightweight CNNs becomes imperative, compelling us to meticulously compare existing lightweight ViT and CNNs.

Lightweight ViTs and lightweight CNNs exhibit certain structural similarities. For example, both of them employ convolutional modules to learn spatially local representations~\cite{mehta2021mobilevit,mehta2022separable,wadekar2022mobilevitv3,pan2022edgevits}. For learning global representations, lightweight CNNs usually enlarge the kernel size of convolutions~\cite{zhang2022parc}, while lightweight ViTs generally employ the multi-head self-attention module~\cite{mehta2021mobilevit,mehta2022separable}.
However, despite these structural connections, there remain notable differences in the block structure, macro/micro designs between them, which have yet to receive sufficient inspection. For example, lightweight ViTs usually adopt the MetaFormer block structure~\cite{yu2022metaformer}, while lightweight CNNs favors the inverted residual bottleneck~\cite{sandler2018mobilenetv2}. This naturally leads us to a question: \textit{Can architectural designs of lightweight ViTs enhance lightweight CNNs' performance?} To answer this question, we revisit the design of lightweight CNNs from the ViT perspective in this study. Our research aims to bridge the gap between lightweight CNNs and lightweight ViTs and highlight the promising prospect of the former for deployment on mobile devices compared to the latter.

To accomplish this objective, following~\cite{liu2022convnet}, we begin with a standard lightweight CNN, \ie, MobileNetV3-L~\cite{howard2019searching}. We gradually ``modernize" its architecture by incorporating the efficient architectural designs of lightweight ViTs~\cite{li2022efficientformer,mehta2021mobilevit,li2022rethinking,liu2023efficientvit}. Finally, for resource-constrained mobile devices, we obtain a new family of lightweight CNNs, namely \textbf{RepViT}, which is composed entirely of \textbf{Re}-\textbf{p}arameterization convolutions in a \textbf{ViT}-like MetaFormer structure~\cite{yu2022metaformer,yu2022metaformerbaseline,li2022efficientformer}. As a pure lightweight CNN, RepViT presents superior performance and efficiency compared with existing state-of-the-art lightweight ViTs~\cite{li2022rethinking,pan2022edgevits} on various computer vision tasks, including image classification on ImageNet~\cite{deng2009imagenet}, object detection and instance segmentation on COCO-2017~\cite{lin2014microsoft}, and semantic segmentation on ADE20k~\cite{zhou2017scene}. Notably, RepViT reaches over 80\% top-1 accuracy on ImageNet, with 1.0 ms latency on an iPhone 12, which is the first time for a lightweight model, to the best of our knowledge. Our largest model, RepViT-M2.3, obtains 83.7\% accuracy with only 2.3 ms latency. After incorporating RepViT with SAM~\cite{kirillov2023segment}, our RepViT-SAM can obtain nearly $10\times$ faster inference speed than the state-of-the-art MobileSAM~\cite{zhang2023faster} while enjoying significantly better zero-shot transfer capability. We hope that RepViT can serve as a strong baseline and inspire further research into lightweight models for edge deployments.

\section{Related Work}
In the past decade, Convolutional Neural Networks (CNNs) have emerged as the predominant approach for computer vision tasks~\cite{he2016deep,he2017mask,wang2023hierarchical,ding2023exploring,lyu2023box,xiong2023confidence} due to their natural inductive locality biases and translation equivalence. However, the extensive computation of standard CNNs renders them unsuitable for deployment on resource-constrained mobile devices. To overcome this challenge, numerous techniques have been proposed to make CNNs more lightweight and mobile-friendly, including separable convolutions~\cite{howard2017mobilenets}, inverted residual bottleneck~\cite{sandler2018mobilenetv2}, channel shuffle~\cite{zhang2018shufflenet,ma2018shufflenet}, 
and structural re-parameterization~\cite{ding2021repvgg}, \etc. These methods have paved the way for the development of several widely used lightweight CNNs, like MobileNets~\cite{howard2017mobilenets,sandler2018mobilenetv2,howard2019searching}, ShuffleNets~\cite{ma2018shufflenet,zhang2018shufflenet},
and RepVGG~\cite{ding2021repvgg}.

Subsequently, the Vision Transformer (ViT)~\cite{dosovitskiy2020image} was introduced, which adapts the transformer architecture to achieve state-of-the-art performance on large-scale image recognition tasks, surpassing that of CNNs~\cite{dosovitskiy2020image,touvron2021training}. Building on the competitive performance of ViTs, subsequent works have sought to incorporate spatial inductive biases to enhance their stability and performance~\cite{guo2022cmt,dai2021coatnet}, design more efficient self-attention operations~\cite{dong2022cswin,zhu2023biformer}, and adapt ViTs to a diverse range of computer vision tasks~\cite{zhang2022topformer,esser2021taming}.


Although ViTs have shown superior performance over CNNs on various vision tasks, most of them are heavy-weighted, requiring substantial computation and memory footprint~\cite{liu2021swin,touvron2021training}. That makes them unsuitable for mobile devices with limited resources~\cite{mehta2021mobilevit,pan2022edgevits}. Consequently, researchers have dedicated to exploring various techniques to make ViTs more lightweight and more friendly for mobile devices~\cite{mehta2022separable,vasu2023fastvit}. For example, MobileViT~\cite{mehta2021mobilevit} adopts a hybrid architecture, combining lightweight MobileNet blocks and multi-head self-attention (MHSA) blocks. EfficientFormer~\cite{li2022efficientformer} proposes a dimension-consistent design paradigm to enhance the latency-performance boundary.
These lightweight ViTs have demonstrated new state-of-the-art performance and latency trade-offs on mobile devices, outperforming previous lightweight CNNs~\cite{vasu2023mobileone,sandler2018mobilenetv2}.

The success of lightweight ViTs is usually attributed to the multi-head self-attention module with the capability of learning global representations. However, the notable architectural distinctions between lightweight CNNs and lightweight ViTs, including their block structures, as well as macro and micro elements, are generally overlooked. As such, distinguished from existing works, our primary goal is to revisit the design of lightweight CNNs by integrating the architectural designs of lightweight ViTs. We aim to bridge the gap between lightweight CNNs and lightweight ViTs, and emphasize the mobile-friendliness of the former.

\section{Methodology}
\begin{figure}[t]
  \centering
   \includegraphics[width=1.0\linewidth]{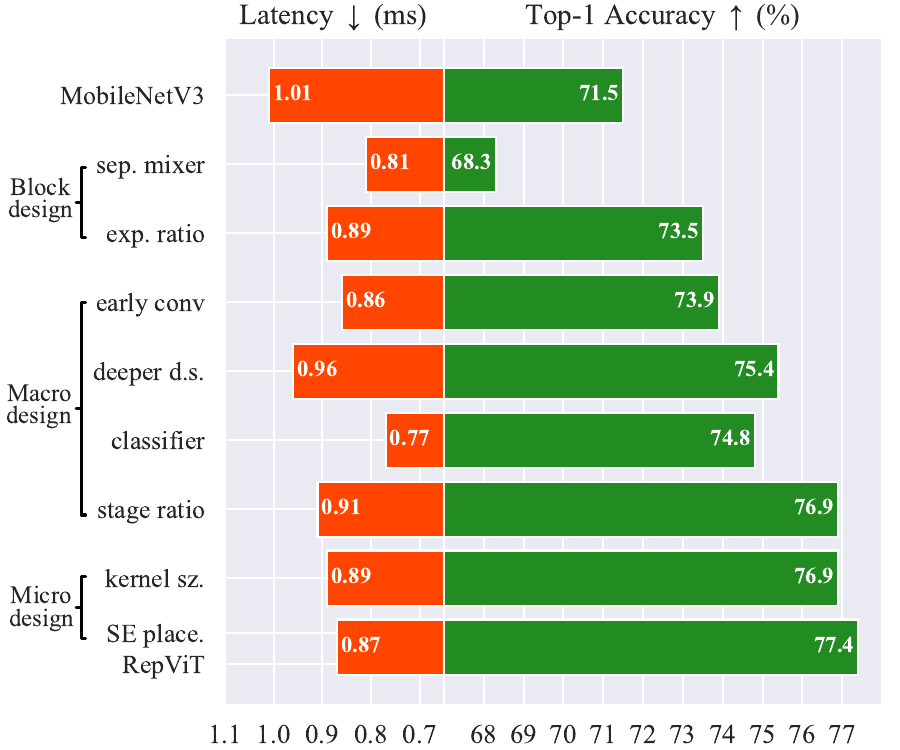}
   \caption{We modernize MobileNetV3-L from various granularities. We mainly consider the latency on mobile devices and the top-1 accuracy on ImageNet-1K. Finally, we obtain a new family of pure lightweight CNNs, namely RepViT, which can achieve lower latency and higher performance. Note that these results are obtained without the distillation.}
   \label{fig:tran}
   \vspace{-10pt}
\end{figure}

In this section, we begin with a standard lightweight CNN, \ie, MobileNetV3-L, and then gradually modernize it from various granularities, by incorporating the architectural designs of lightweight ViTs. We first introduce the metric to measure the latency on mobile devices, and then align the training recipe with existing lightweight ViTs in \Cref{sec:preliminary}. Based on the consistent training setting, we explore the optimal block design in \Cref{sec:block}. We further optimize the performance of MobileNetV3-L on mobile devices from macro-architectural elements in \Cref{sec:macro}, \ie, stem, downsampling layers, classifier and overall stage ratio. We then tune the lightweight CNN through layer-wise micro designs in \Cref{sec:micro}. \Cref{fig:tran} shows the whole procedure and results we achieve in each step. Finally, we obtain a new family of pure lightweight CNNs designed for mobile devices in \Cref{sec:network}, namely RepViT. All models are trained and evaluated on ImageNet-1K.

\subsection{Preliminary}
\label{sec:preliminary}

\paragraph{Latency metric.}
Previous works~\cite{chen2022mobile,tan2019efficientnet} optimize the inference speed of models based on metrics like floating point operations (FLOPs) or model sizes. However, these metrics do not correlate well with real-world latency in mobile applications~\cite{li2022efficientformer}. Hence, following \cite{vasu2023mobileone,li2022efficientformer,li2022rethinking,mehta2021mobilevit}, we measure the actual on-device latency as the benchmark metric. Such a strategy can provide a more accurate performance evaluation and fair comparisons among different models on real-world mobile devices. In practice, we utilize the iPhone 12 as the test device and Core ML Tools~\cite{coreml} as the compiler, like~\cite{vasu2023mobileone,li2022efficientformer,li2022rethinking}. Besides, to avoid unsupported functions with Core ML Tools, we employ the GeLU activation in the MobileNetV3-L model, following \cite{li2022efficientformer,vasu2023mobileone}.

\textit{We measure the latency of MobileNetV3-L to be 1.01 ms.}

\paragraph{Aligning training recipe.}
Recent lightweight ViTs~\cite{li2022efficientformer,li2022rethinking,pan2022edgevits,mehta2021mobilevit} generally adopt the training recipe from DeiT~\cite{touvron2021training}. Specifically, they use the AdamW optimizer~\cite{loshchilov2017decoupled} and the cosine learning rate scheduler to train the models from scratch for 300 epochs, with a teacher of RegNetY-16GF~\cite{radosavovic2020designing} for distillation. Besides, they adopt Mixup~\cite{zhang2017mixup}, auto-augmentation~\cite{cubuk2019autoaugment}, and random erasing~\cite{zhong2020random} for data augmentation. Label Smoothing~\cite{szegedy2016rethinking} is also employed as the regularization scheme. For fair comparisons, we align the training recipe of MobileNetV3-L with the existing lightweight ViTs, with the exception of excluding knowledge distillation for now. Consequently, MobileNetV3-L obtains 71.5\% top-1 accuracy.

\textit{We will now use this training recipe by default.}

\subsection{Block design}
\label{sec:block}

\paragraph{Separate token mixer and channel mixer.}
The block structure of lightweight ViTs~\cite{li2022efficientformer,li2022rethinking,mehta2022separable} incorporates an important design feature, namely the separate token mixer and channel mixer~\cite{yu2022metaformerbaseline}. According to the recent research~\cite{yu2022metaformer}, the effectiveness of ViTs primarily originates from their general token mixer and channel mixer architecture, \ie, the MetaFormer architecture, rather than the equipped specific token mixer. In light of this finding, we aim to emulate the existing lightweight ViTs by splitting the token mixer and channel mixer in MobileNetV3-L. 

\begin{figure}[t]
\centering
    \includegraphics[width=1.0\linewidth]{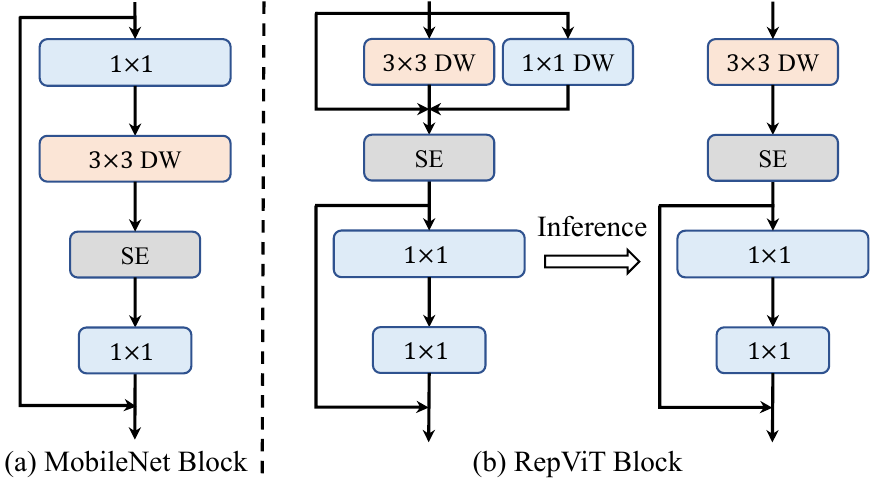}
    \caption{\textbf{Block design}. (a) is a MobileNetV3 block with an optional squeeze-and-excitation (SE) layer. (b) is the designed RepViT block, which separates the token mixer and channel mixer through the structural re-parameterization technique. The SE layer is also optional in RepViT block. The norm layer and nonlinearity are omitted for simplicity.}
    \label{fig:block}
    \vspace{-15pt}
\end{figure}

Specifically, as depicted in ~\Cref{fig:block}.(a), the original MobileNetV3 block adopts a $1\times1$ expansion convolution, and a $1\times1$ projection layer to enable interaction among channels (\ie, channel mixer). A $3\times3$ depthwise (DW) convolution is equipped after the $1\times1$ expansion convolution for the fusion of spatial information (\ie, token mixer). Such a design makes the token mixer and channel mixer coupled together. In order to separate them, we first move up the DW convolution. The optional squeeze-and-excitation (SE) layer is also moved up to be placed after the DW, as it depends on spatial information interaction. Consequently, we can successfully separate the token mixer and channel mixer in the MobileNetV3 block. We further employ a widely used structural re-parameterization technique~\cite{chu2022make, ding2021repvgg} for the DW layer to enhance the model learning during training. Thanks to the structural re-parameterization technique, we can eliminate the computational and memory costs associated with the skip connection during inference, which is especially advantageous for mobile devices. We name such a block as RepViT block (\Cref{fig:block}.(b)), which reduces the latency of MobileNetV3-L to 0.81 ms, together with a temporary performance degradation to 68.3\%.


\paragraph{Reducing expansion ratio and increasing width.}
\label{paragraph:exp}
In vanilla ViTs, the expansion ratio in the channel mixer is typically set to 4, making the hidden dimension of the Feed Forward Network (FFN) module $4\times$ wider than the input dimension. It thus consumes a significant portion of the computation resource, thereby contributing substantially to the overall inference time~\cite{zheng2022savit}. To alleviate this bottleneck, recent works~\cite{jiang2021all,graham2021levit} employ a narrower FFN. For instance, LV-ViT~\cite{jiang2021all} adopts an expansion ratio of 3 in FFN. LeViT~\cite{graham2021levit} sets the expansion ratio to 2. Besides, Yang \etal~\cite{yang2021nvit} point out that there exists a significant amount of channel redundancy in FFN. Therefore, it is reasonable to use a smaller expansion ratio. 

In MobileNetV3-L, the expansion ratio ranges from 2.3 to 6, with a concentration of 6 in the last two stages that have a greater number of channels. For our RepViT block, we set the expansion ratio to 2 in the channel mixer for all stages, following~\cite{jiang2021all,liu2023efficientvit,graham2021levit}. This results in a latency reduction to 0.65 ms. Consequently, with the smaller expansion ratio, we can increase the network width to remedy the large parameter reduction. We double the channels after each stage, ending up with 48, 96, 192, and 384 channels for each stage, respectively. 
These modifications can increase the top-1 accuracy to 73.5\% with a latency of 0.89 ms.


Note that, by directly adjusting the expansion ratio and network width on the original MobileNetV3 block, we obtain inferior performance with 73.0\% top-1 accuracy under a similar latency of 0.91 ms. Therefore, by default, for the block design, \textit{we employ the new expansion ratio and network width with the RepViT block.}


\begin{figure*}[t]
\centering
    \includegraphics[width=0.95\linewidth]{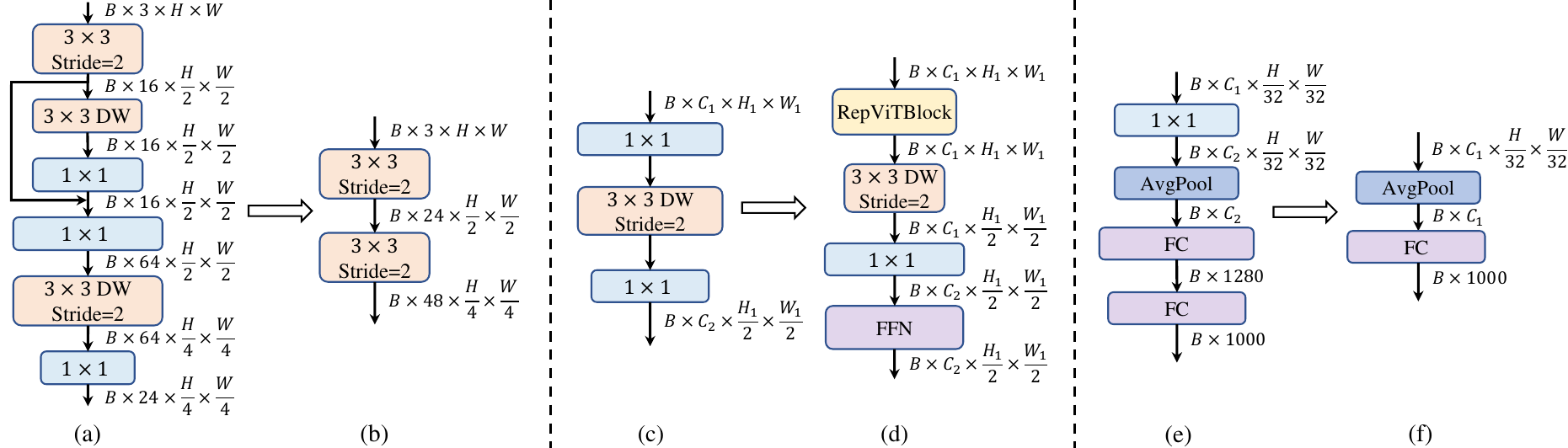}
    \caption{\textbf{Macro design}. (a) and (b), (c) and (d), (e) and (f) indicate the designs for stem, downsampling layer and classifier, respectively. RepViT has four stages with $\frac{H}{4}\times \frac{W}{4}$, $\frac{H}{8}\times \frac{W}{8}$, $\frac{H}{16}\times \frac{W}{16}$ and $\frac{H}{32}\times \frac{W}{32}$ resolutions respectively, where $H$ and $W$ denote the width and height of the input image. $C$ represents the channel dimension and $B$ denotes the batch size. The norm layer and nonlinearity are omitted.}
    \label{fig:macro}
    \vspace{-10pt}
\end{figure*}

\subsection{Macro design}
\label{sec:macro}
In this part, we carry out optimizations with a specific focus on its macro architecture for mobile devices, from the front to the back of the network.

\paragraph{Early convolutions for stem.} ViTs typically use a patchify operation as the stem, dividing the input image into non-overlapping patches~\cite{dosovitskiy2020image}. This simple stem corresponds to a non-overlapping convolution with a large kernel size (\eg, kernel size = 16) and a large stride (\eg, stride = 16). Hierarchical ViTs~\cite{liu2021swin,wang2021pyramid} adopt the same patchify operation, but with a smaller patch size of 4. However, recent work~\cite{xiao2021early} shows that such a patchify operation easily causes the substandard optimizability and sensitivity to training recipes for ViTs. To mitigate these problems, they suggest using a small number of stacked stride-two 3$\times$3 convolutions as an alternative for the stem, known as early convolutions, which improves the optimization stability and performance, and is thus widely adopted by lightweight ViTs~\cite{li2022efficientformer,li2022rethinking}. 

In contrast, MobileNetV3-L adopts a complex stem that involves a $3\times3$ convolution, a depthwise separable convolution, and an inverted bottleneck, as shown in \Cref{fig:macro}.(a). Since the stem module processes the input image at the highest resolution, a complex architecture can suffer from severe latency bottlenecks on mobile devices. Therefore, as a trade-off, MobileNetV3-L reduces the initial number of filters to 16, which in turn limits the representation power of the stem. To address these issues, following \cite{xiao2021early,liu2023efficientvit,li2022efficientformer,li2022rethinking}, we employ the way of early convolutions~\cite{xiao2021early} and simply equip two $3\times3$ convolutions with stride = 2 as the stem. As shown in \Cref{fig:macro}.(b), the number of filters in the first convolution is set to 24 and the one in the second is set to 48. The overall latency is reduced to 0.86 ms. Meanwhile, the top-1 accuracy is improved to 73.9\%.

\textit{We will now use early convolutions as the stem.}


\paragraph{Deeper downsampling layers.} In ViTs, spatial downsampling is typically achieved by a separate patch merging layer. As demonstrated in \cite{liu2022convnet}, such a separation-based downsampling layer facilitates an increase in network depth and mitigates the information loss due to the resolution reduction. Therefore, EfficientViT~\cite{liu2023efficientvit} adopts a sandwich layout to deepen the downsampling layer, achieving efficient and effective downsampling. In contrast, MobileNetV3-L achieves downsampling only by an inverted bottleneck block with the DW convolution of stride = 2, as illustrated in \Cref{fig:macro}.(c). This design may lack adequate network depth, resulting in information loss and negative impact on the model performance. Therefore, to achieve a separate and deeper downsampling layer, we first use a DW convolution with stride = 2 and a pointwise $1\times1$ convolution to perform the spatial downsampling and modulate the channel dimension, respectively, as shown in \Cref{fig:macro}.(d). Besides, we prepend a RepViT block to further deepen the downsampling layer. A FFN module is placed after the $1\times1$ convolution to memorize more latent information. As a result, such deeper downsampling layers bring the top-1 accuracy to 75.4\% with a latency of 0.96 ms.


\textit{We will now leverage the deeper downsampling layers.}


\paragraph{Simple classifier.} In lightweight ViTs~\cite{li2022efficientformer,mehta2021mobilevit,graham2021levit}, the classifier generally consists of a global average pooling layer followed by a linear layer. Such a simple classifier is thus friendly to the latency, especially for mobile devices. In contrast, MobileNetV3-L employs a complicated classifier, which includes one extra $1\times1$ convolution and one extra linear layer to expand the features to a higher-dimensional space~\cite{cui2021pp}, as shown in \Cref{fig:macro}.(e). Such a design is crucial for MobileNetV3-L to generate rich predictive features~\cite{howard2019searching}, particularly given the small output channel in the final stage. However, it in turn results in a heavy burden to the latency on mobile devices. Considering that the final stage now has more channels after block design in \Cref{paragraph:exp}, we thus replace it with a simple classifier, \ie, a global average pooling layer and a linear layer, as shown in \Cref{fig:macro}.(f). This step causes an accuracy drop of 0.6\% but make the latency decrease to 0.77 ms.

\textit{We will now employ the simple classifier.}


\paragraph{Overall stage ratio.} Stage ratio represents the ratio of the number of blocks in different stages, thereby indicating the distribution of computation across the stages. Previous works~\cite{radosavovic2019network,radosavovic2020designing} have shown that the utilization of more blocks in the third stage confers a favorable balance between the accuracy and speed. Therefore, existing lightweight ViTs generally apply more blocks in this stage. For example, EfficientFormer-L2~\cite{li2022efficientformer} employs a stage ratio of 1:1:3:1.5. Meanwhile, Conv2Former~\cite{hou2022conv2former} shows that a more aggressive stage ratio and a deeper layout perform better for small models. They thus adopt the stage ratio of 1:1:4:1 and 1:1:8:1 for Conv2Former-T and Conv2Former-S, respectively. Here, we employ a stage ratio of 1:1:7:1 for the network. We then increase the network depth to 2:2:14:2, achieving a deeper layout. This step increases the top-1 accuracy to 76.9\% with a latency of 0.91 ms.

\textit{We will use this stage ratio.}

\begin{table*}[th]
\caption{{\textbf{Classification performance on ImageNet-1K.} Following~\cite{graham2021levit,liu2023efficientvit}, throughput is tested on a Nvidia RTX3090 GPU with maximum power-of-two batch size that fits in memory.}}
\label{tab:comparison}
\centering
\small
\scalebox{0.95}{
\begin{tabular}{cccccccccc}
\toprule
\multirow{2}{*}{Model}           & \multirow{2}{*}{Type}      & \multirow{2}{*}{Params (M)} & \multirow{2}{*}{GMACs} &  \multirow{2}{*}{\makecell{Latency $\downarrow$ \\ (ms)}} & \multirow{2}{*}{\makecell{Throughput $\uparrow$ \\ (im/s)}} & \multirow{2}{*}{Epochs} & \multirow{2}{*}{Top-1 (\%)}  \\
\\
\hline
MobileViG-Ti~\cite{munir2023mobilevig}  &   CNN-GNN   & 5.2      &  0.7    & 1.0      & 4337 & 300  & 75.7    \\
FastViT-T8~\cite{vasu2023fastvit} & Hybrid & 3.6 & 0.7 & 0.9 & 3909 & 300 & 76.7 \\
SwiftFormer-XS~\cite{shaker2023swiftformer} & Hybrid & 3.5 & 0.6 & 1.0 & 4304 & 300 & 75.7 \\
EfficientFormerV2-S0~\cite{li2022rethinking}  &   Hybrid   & 3.5      &  0.4    & 0.9      & 1274 & 300 / 450 & 75.7 / 76.2   \\
\rowcolor[gray]{0.92}
\textbf{RepViT-M0.9} & CONV & 5.1 & 0.8 & \textbf{0.9} & 4817 & 300 / 450 & \textbf{78.7} / \textbf{79.1} \\
\rowcolor[gray]{0.92}
\textbf{RepViT-M1.0} & CONV & 6.8 & 1.1 & \textbf{1.0} & 3910 & 300 / 450 & \textbf{80.0} / \textbf{80.3} \\
\hline
MobileViG-S~\cite{munir2023mobilevig}  &   CNN-GNN   & 7.2      &  1.0    & 1.2      & 2985 & 300  & 78.2    \\
EfficientFormer-L1~\cite{li2022efficientformer}       & Hybrid    &    12.3   & 1.3    & 1.4  & 3360 & 300 & 79.2    \\
SwiftFormer-S~\cite{shaker2023swiftformer} & Hybrid & 6.1 & 1.0 & 1.2 & 3376 & 300 & 78.5 \\
EfficientFormerV2-S1~\cite{li2022rethinking}  &   Hybrid   & 6.1      &  0.7    & 1.1     & 1153 & 300 / 450 & 79.0 / 79.7  \\
\rowcolor[gray]{0.92}
\textbf{RepViT-M1.1} & CONV & 8.2 & 1.3 & \textbf{1.1} & 3604 & 300 / 450 & \textbf{80.7} / \textbf{81.2} \\
\hline
MobileViG-M~\cite{munir2023mobilevig}  &   CNN-GNN   & 14.0      &  1.5    & 1.6      & 2491 & 300  & 80.6    \\
FastViT-S12~\cite{vasu2023fastvit} & Hybrid & 8.8 & 1.8 & 1.5 & 2313 & 300 & 80.9 \\
FastViT-SA12~\cite{vasu2023fastvit} & Hybrid & 10.9 & 1.9 & 1.8 & 2181 & 300 & 81.9 \\
SwiftFormer-L1~\cite{shaker2023swiftformer} & Hybrid & 12.1 & 1.6 & 1.6 & 2576 & 300 & 80.9 \\
EfficientFormerV2-S2~\cite{li2022rethinking}  &   Hybrid   &  12.6  &  1.3    & 1.6     & 611 & 300 / 450 & 81.6 / 82.0    \\
\rowcolor[gray]{0.92}
\textbf{RepViT-M1.5} & CONV & 14.0 & 2.3 & \textbf{1.5} & 2151 & 300 / 450 & \textbf{82.3} / \textbf{82.5} \\
\hline
MobileViG-B~\cite{munir2023mobilevig} & CNN-GNN & 26.7 & 2.8 & 2.7 & 1446 & 300 & 82.6 \\
EfficientFormer-L3~\cite{li2022efficientformer}  &   Hybrid   & 31.3   &  3.9    & 2.7    & 1422 & 300 & 82.4  \\
EfficientFormer-L7~\cite{li2022efficientformer}  &   Hybrid   & 82.1   &  10.2    & 6.6    & 619 & 300 & 83.3  \\
SwiftFormer-L3~\cite{shaker2023swiftformer} & Hybrid & 28.5 & 4.0 & 2.9 & 1474 & 300 & 83.0 \\
EfficientFormerV2-L~\cite{li2022rethinking}  &   Hybrid   & 26.1   &  2.6    & 2.7     & 399 & 300 / 450 & 83.3 / 83.5  \\
\rowcolor[gray]{0.92}
\textbf{RepViT-M2.3} & CONV & 22.9 & 4.5 & \textbf{2.3} & 1184 & 300 / 450 & \textbf{83.3} / \textbf{83.7} \\
\bottomrule
\end{tabular}
}
\vspace{-5pt}
\end{table*}

\subsection{Micro design}
\label{sec:micro}
In this section, we focus on the micro architecture for lightweight CNNs, including the kernel size selection and squeeze-and-excitation (SE) layer placement.

\paragraph{Kernel size selection.} The performance and latency of CNNs are often impacted by the size of convolution kernels. For example, to capture long-range dependencies like MHSA, ConvNeXt~\cite{liu2022convnet} employs large kernel-sized convolutions, exhibiting the performance gain. Similarly, RepLKNet~\cite{ding2022scaling} shows a powerful paradigm that utilizes super large convolution kernels in CNNs. However, large kernel-sized convolution is not friendly for mobile devices, due to its computation complexity and memory access costs. Additionally, compared to $3\times3$ convolutions, larger convolution kernels are typically not highly optimized by compilers and computing libraries~\cite{ding2021repvgg}. MobileNetV3-L primarily utilizes $3\times3$ convolutions, with a small number of $5\times5$ convolutions employed in certain blocks. To ensure the inference efficiency on the mobile device, we prioritize the simple $3\times3$ convolutions in all modules. This replacement can maintain the top-1 accuracy at 76.9\% while enjoying a latency reduction to 0.89 ms.

\textit{We will now use $3\times3$  convolutions.}

\paragraph{Squeeze-and-excitation layer placement.} One advantage of self-attention module compared with convolution is the ability to adapt weights according to input, known as the data-driven attribute~\cite{jaderberg2015spatial,woo2018cbam}. As a channel wise attention module, SE layers~\cite{hu2018squeeze} can compensate for the limitation of convolutions in lacking data-driven attributes, bringing better performance~\cite{zhang2022parc}. MobileNetV3-L incorporates SE layers in certain blocks, with a primary focus on the latter two stages. However, as shown in~\cite{ridnik2021tresnet}, stages with low-resolution feature maps get a smaller accuracy benefit, compared to stages with higher resolution feature maps. Meanwhile, along with performance gains, SE layers also introduce non-negligible computational costs. Therefore, we design a strategy to utilize SE layers in a cross-block manner, \ie, adopting the SE layer in the 1st, 3rd, 5th, ... block in each stage, to maximize the accuracy benefit with a minimal latency increment. This step brings the top-1 accuracy to 77.4\% with a latency of 0.87 ms.

\textit{We will now use this cross-block SE layer placement. This brings our final model, namely RepViT.}

\subsection{Network architecture}
\label{sec:network}

Following~\cite{li2022efficientformer,mehta2021mobilevit}, we develop multiple RepViT variants, including RepViT-M0.9/M1.0/M1.1/M1.5/M2.3. The suffix "-MX" means that the latency of the corresponding model is X ms on the mobile device, \ie, iPhone 12 with iOS 16. Variants are distinguished by the number of channels and the number of blocks within each stage. Please refer to the supplementary material for more details.

\section{Experiments}

\subsection{Image Classification}

\paragraph{Implementation details.} We conduct image classification experiments on ImageNet-1K, using a standard image size of 224$\times$224 for both training and testing. Following \cite{vasu2023fastvit,li2022efficientformer,li2022rethinking,munir2023mobilevig}, we train all models from scratch for 300 epochs or 450 epochs using the same training recipe.
For fair comparisons, the RegNetY-16GF model with a top-1 accuracy of 82.9\% is used as the teacher model for distillation. Following~\cite{li2022efficientformer,li2022rethinking,vasu2023mobileone}, the latency is measured on iPhone 12 with models compiled by Core ML Tools under a batch size of 1. Note that RepViT-M0.9 is the outcome of the ``modernizing" process applied to MobileNetV3-L. Following~\cite{vasu2023fastvit,li2022rethinking}, we report the performance with and without distillation in \Cref{tab:comparison} and \Cref{tab:distillation}, respectively.

\begin{table}[t]
  \caption{Results without distillation on ImageNet-1K.}
  \label{tab:distillation}
  \centering
  \small
  \scalebox{0.95}{
  \begin{tabular}{cccccccccc}
  \toprule
  \multirow{1}{*}{Model}          &  \multirow{1}{*}{\makecell{Latency (ms)}} & \multirow{1}{*}{Epochs} & \multirow{1}{*}{Top-1 (\%)} \\
  \hline
  MobileOne-S1~\cite{vasu2023mobileone} & 0.9 & 300 & 75.9 \\
  EfficientFormerV2-S0~\cite{li2022rethinking} & 0.9 & 300 & 73.7 \\
  FastViT-T8~\cite{vasu2023fastvit} & 0.9 & 300 & 75.6 \\
  \rowcolor[gray]{0.92}
  \textbf{RepViT-M0.9} & \textbf{0.9} & 300 & \textbf{77.4} \\
  \rowcolor[gray]{0.92}
  \textbf{RepViT-M1.0} & \textbf{1.0} & 300 & \textbf{78.6} \\
  \hline
  MobileOne-S2~\cite{vasu2023mobileone} & 1.1 & 300 & 77.4 \\
  EdgeViT-XS~\cite{pan2022edgevits} & 3.6 & 300 & 77.5 \\
  EfficientFormerV2-S1~\cite{li2022rethinking} & 1.1 & 300 & 77.9 \\
  \rowcolor[gray]{0.92}
  \textbf{RepViT-M1.1} & \textbf{1.1} & 300 & \textbf{79.4} \\
  \hline
  \rowcolor[gray]{0.92}
  MobileOne-S4~\cite{vasu2023mobileone} & 1.6 & 300 & 79.4 \\
  FastViT-S12~\cite{vasu2023fastvit} & 1.5 & 300 & 79.8 \\
  EfficientFormerV2-S2~\cite{li2022rethinking} & 1.6 & 300 & 80.4 \\
  \textbf{RepViT-M1.5} & \textbf{1.5} & 300 & \textbf{81.2} \\
  \hline
  EfficientNet-B3~\cite{tan2019efficientnet} & 5.3 & 350 & 81.6 \\
  PoolFormer-S36~\cite{yu2022metaformer} & 3.5 & 300 & 81.4 \\
  \rowcolor[gray]{0.92}
  \textbf{RepViT-M2.3} & \textbf{2.3} & 300 & \textbf{82.5} \\
  \bottomrule
  \end{tabular}
  }
  \vspace{-10pt}
\end{table}

\paragraph{Comparison with state-of-the-arts.} As shown in \Cref{tab:comparison}, RepViT consistently achieves state-of-the-art performance across various model sizes. 
With similar latency, RepViT-M0.9 can significantly outperform EfficientFormerV2-S0 and FastViT-T8 by 3.0\% and 2.0\% top-1 accuracy, respectively. RepViT-M1.1 can also enjoy 1.7\% performance improvement over EfficientFormerV2-S1. It is worth noting that RepViT-M1.0 notably achieves over 80\% top-1 accuracy with 1.0 ms latency on iPhone 12, which is the first time for a lightweight model, to the best of our knowledge. Our largest model, RepViT-M2.3, obtains 83.7\% accuracy with only 2.3 ms latency. The results above well demonstrate that pure lightweight CNNs can outperform existing the state-of-the-art lightweight ViTs on mobile devices by incorporating the efficient architectural designs.

\paragraph{Results without knowledge distillation.} As shown in \Cref{tab:distillation}, even without the enhancement of knowledge distillation, our RepViT can still significantly outperform all competitor models in different levels of latency. For example, with a latency of 1.0 ms, our RepViT-M1.0 can enjoy 2.7\% accuracy gain over MobileOne-S1. For larger models, our RepViT-M2.3 can obtain 1.1\% performance improvement while enjoying 34.3\% latency reduction (3.5 ms to 2.3 ms), compared with PoolFormer-S36. Such results further demonstrate the effectiveness of our models.

\subsection{RepViT meets SAM}
Segment Anything Model (SAM)~\cite{kirillov2023segment} has shown impressive zero-shot transfer performance for various computer vision tasks recently. However, its heavy computation costs remain daunting for resource-constrained mobile devices. Here, to show the promising performance of RepViT in segmenting anything on mobile devices, following~\cite{zhang2023faster}, we replace the heavyweight image encoder in SAM with our RepViT model, ending up with the RepViT-SAM model. RepViT-SAM employs RepViT-M2.3 as the image encoder and is trained for 8 epochs under the same setting as~\cite{zhang2023faster}. Like MobileSAM~\cite{zhang2023faster}, we use only 1\% data in the SAM-1B dataset~\cite{kirillov2023segment} for training. The project page can be found at \url{https://jameslahm.github.io/repvit-sam/}.

\begin{table}[t]
  \centering
  \caption{
    Comparison between RepViT-SAM and others in terms of latency. The latency (ms) is measured with the standard resolution~\cite{ghiasi2021simple} of 1024$\times$1024 on iPhone 12 and Macbook M1 Pro by Core ML Tools. OOM means out of memory.
  }
  \small
  \resizebox{\linewidth}{!}{
  \begin{tabular}{c|c|c|c|c}
  \toprule
   \multirow{2}{*}{Platform} &  \multicolumn{3}{c|}{Image encoder}  & \multirow{2}{*}{\shortstack{Mask\\ decoder}} \\
    \cmidrule{2-4}
    & \textbf{RepViT-SAM} & MobileSAM~\cite{zhang2023faster} & ViT-B-SAM~\cite{kirillov2023segment} \\
    \midrule
    iPhone & \textbf{48.9} & OOM & OOM & 11.6\\
    Macbook & \textbf{44.8} & 482.2 & 6249.5 & 11.8 \\
  \bottomrule
  \end{tabular}
  }
  \label{tab:sam-latency}
  \vspace{-5pt}
\end{table}

\begin{table}[t]
\centering
\caption{
    Comparison results on zero-shot edge detection (z.s. edge.), zero-shot instance segmentation (z.s. ins.), and segmentation in the wild benchmark (SegInW). Bold indicates the best, and underline indicates the second best.
}
\small
\resizebox{\linewidth}{!}{
\begin{tabular}{c|ccc|c|c}
\toprule
\multirow{2}{*}{Model} & \multicolumn{3}{c|}{z.s. edge.} & z.s. ins. & SegInW \\
    \cmidrule{2-6} 
    & ODS & OIS & AP & AP & Mean AP\\
    \midrule
    ViT-H-SAM~\cite{kirillov2023segment}  & \textbf{.768} & \textbf{.786} & \textbf{.794} & \textbf{46.8} & \textbf{48.7} \\
    ViT-B-SAM~\cite{kirillov2023segment} & .743 & .764 & .726 & 42.5 & 44.8 \\
    MobileSAM~\cite{zhang2023faster} & .756 & .768 & .746 & 42.7 & 43.9 \\
    \rowcolor[gray]{0.92}
    RepViT-SAM & \underline{.764} & \textbf{.786} & \underline{.773} & \underline{44.4} & \underline{46.1} \\
\bottomrule
\end{tabular}
}
\label{tab:sam}
\vspace{-10pt}
\end{table}

We first compare our RepViT-SAM with MobileSAM~\cite{zhang2023faster} and the original SAM~\cite{kirillov2023segment} with ViT-B image encoder, \ie, ViT-B-SAM, in terms of latency. As demonstrated in \Cref{tab:sam-latency}, on iPhone 12, our RepViT-SAM can perform model inference smoothly, while both competitors fail to run. On Macbook M1 Pro, RepViT-SAM is nearly 10$\times$ faster than the state-of-the-art MobileSAM. 

We then evaluate the performance of our RepViT-SAM on zero-shot edge detection using BSDS500~\cite{martin2001database,arbelaez2010contour}, zero-shot instance segmentation using COCO~\cite{lin2014microsoft}, and segmentation in the wild benchmark (SegInW), following~\cite{kirillov2023segment,ke2023segment}. As shown in~\Cref{tab:sam}, our RepViT-SAM outperforms MobileSAM and ViT-B-SAM on all benchmarks. Compared with ViT-H-SAM, which is the largest SAM model with over 615M parameters, our small RepViT-SAM can obtain comparable performance in terms of ODS and OIS on the zero-shot edge detection. Overall, taking all the results into consideration, our RepViT-SAM model exhibits exceptional efficiency on both the iPhone 12 and Macbook M1 Pro, while maintaining remarkable transfer performance for downstream tasks. We hope that RepViT-SAM model can serve as a strong baseline for SAM on edge deployments.

\begin{table*}[h]
  \centering
  \small
  \caption{
  \textbf{Object detection \&
  instance segmentation} results on MS COCO 2017 with the Mask RCNN framework.
  \textbf{Semantic segmentation} results on ADE20K by integrating models into Semantic FPN. Backbone latencies are measured with image crops of 512$\times$512 on iPhone 12 by Core ML Tools. * indicates that the model is initialized with weights pretrained for 450 epochs on ImageNet-1K.
  }
  \resizebox{0.9\linewidth}{!}{
  \begin{tabular}{c|c|ccc|ccc|c}
  \toprule
  \multirow{2}{*}{Backbone} & \multirow{2}{*}{\makecell{Latency $\downarrow$ \\ (ms)}} & \multicolumn{3}{c|}{Object Detection} & \multicolumn{3}{c|}{Instance Segmentation} & \multicolumn{1}{c}{Semantic}  \\ \cline{3-9}
                            &                         & AP$^{box}$    & AP$^{box}_{50}$   & AP$^{box}_{75}$   & AP$^{mask}$    & AP$^{mask}_{50}$   & AP$^{mask}_{75}$   & mIoU   \\
                            \hline
                            \hline
  ResNet18~\cite{he2016deep}                  &     4.4        & 34.0   & 54.0    & 36.7    & 31.2   & 51.0    & 32.7   & 32.9   \\
  PoolFormer-S12~\cite{yu2022metaformer}            &     7.5          & 37.3   & 59.0    & 40.1    & 34.6   & 55.8    & 36.9     & 37.2  \\
  EfficientFormer-L1~\cite{li2022efficientformer}        &     5.4              & 37.9   & 60.3    & 41.0    & 35.4   & 57.3    & 37.3   &  38.9  \\
  \rowcolor[gray]{0.92}
  RepViT-M1.1     &     \text{4.9}              &  \text{39.8}  &  \text{61.9}   & \text{43.5}  &    \text{37.2}    &   \text{58.8}      &  \text{40.1}   & \text{40.6}  \\
  PoolFormer-S24~\cite{yu2022metaformer}            &     12.3            & 40.1   & 62.2    & 43.4    & 37.0   & 59.1    & 39.6  &  40.3 \\
  PVT-Small~\cite{wang2021pyramid} & 53.7  & 40.4 & 62.9 & 43.8 & 37.8 & 60.1 & 40.3 & 39.8 \\
  EfficientFormer-L3~\cite{li2022efficientformer}        &         12.4     & 41.4   & \text{63.9}    & 44.7    & 38.1   & \text{61.0}    & 40.4   &  43.5 \\
  \rowcolor[gray]{0.92}
  RepViT-M1.5      &       \text{6.4}      & \text{41.6}   &  63.2  &  \text{45.3}   &  \text{38.6}   &  60.5   & \text{41.5}     &   \text{43.6}   \\
  EfficientFormerV2-S2*~\cite{li2022rethinking}        &         12.0     & 43.4   & 65.4    & 47.5    & 39.5   & 62.4    & 42.2   &  42.4 \\
  EfficientFormerV2-L*~\cite{li2022rethinking}        &        18.2     & \text{44.7}   & \text{66.3}    & \text{48.8}   & 40.4   & 63.5   & 43.2   &  45.2 \\
  \rowcolor[gray]{0.92}
  RepViT-M2.3*      &       \text{9.9}      & 44.6   &  66.1  &  \text{48.8}   &  \text{40.8}   &  \text{63.6}  & \text{43.9}     &   \text{46.1}   \\
  \bottomrule
  \end{tabular}
  }
  \vspace{-5pt}
  \label{tab:coco}
\end{table*}

\subsection{Downstream Tasks}

\paragraph{Object Detection and Instance Segmentation.}
We evaluate RepViT on object detection and instance
segmentation tasks to verify its transfer ability. Following \cite{li2022rethinking}, we integrate RepViT into the Mask-RCNN framework~\cite{he2017mask} and conduct experiments on MS COCO 2017~\cite{lin2014microsoft}. 
As seen in \Cref{tab:coco}, RepViT consistently outperforms the competitor models in terms of latency, AP$^{box}$ and AP$^{mask}$, under similar model sizes. Specifically, RepViT-M1.1 significantly outperforms EfficientFormer-L1 backbone by 1.9 AP$^{box}$ and 1.8 AP$^{mask}$, with a smaller latency. For a larger model size, RepViT-M1.5 surpasses EfficientFormer-L3 with a nearly 2$\times$ faster speed while enjoying comparable performance. Compared with EfficientFormerV2-L, RepViT-M2.3 achieves comparable AP$^{box}$ and higher AP$^{mask}$ with a nearly 50\% latency, highlighting the substantial advantage of lightweight CNNs in high-resolution vision tasks. The results above well demonstrate the superiority of RepViT in transferring to downstream vision tasks.

\paragraph{Semantic Segmentation.}
We conduct experiments on ADE20K~\cite{zhou2017scene} to verify the performance of RepViT on the semantic segmentation task. Following \cite{li2022efficientformer,li2022rethinking}, we integrate RepViT into the Semantic FPN framework~\cite{kirillov2019panoptic}. 
As shown in \Cref{tab:coco}, RepViT shows favorable mIoU-latency trade-offs across different model sizes. Specifically, RepViT-M1.1 significantly outperforms EfficientFormer-L1 by 1.7 mIoU with a faster speed. RepViT-M1.5 achieves a 1.2 higher mIoU over EfficientFormerV2-S2, along with a nearly 50\% latency reduction. Compared with EfficientFormerV2-L, RepViT-M2.3 presents an increase of 0.9 mIoU while being nearly 2$\times$ faster. All results show the efficacy of RepViT as a general vision backbone.

\subsection{Model Analyses}

\paragraph{Structural re-parameterization (SR).} To verify the effectiveness of SR in RepViT block, we conduct ablation studies on ImageNet-1K by removing the multi-branch topology of SR at training time. As shown in \Cref{tab:rep}, without SR, different variants of the proposed RepViT suffer from consistent performance declines. The results well demonstrate the positive impact of SR.

\begin{table}[t]
  \centering
  \caption{
    Analyses on structural re-parameterization (SR).
  }
  \small
  \begin{tabular}{c|c|c|c}
  \toprule
   SR & RepViT-M0.9 & RepViT-M1.5 & RepViT-M2.3 \\
   \hline
   $\times$ & 78.47\% & 82.09\% & 83.10\% \\
   \checkmark & 78.74\% & 82.29\% & 83.30\% \\
  \bottomrule
  \end{tabular}
  \vspace{-5pt}
  \label{tab:rep}
\end{table}

\begin{table}[t]
\centering
\caption{
Analyses on SE layer placement.
}
\small
\resizebox{\linewidth}{!}{
\begin{tabular}{c|c|c|c|c}
\toprule
    \multirow{2}{*}{SE} & \multicolumn{2}{c|}{RepViT-M0.9} & \multicolumn{2}{c}{RepViT-M1.5} \\
    \cline{2-5}
    & Top-1 & Latency $\downarrow$ & Top-1 & Latency $\downarrow$\\
    \hline
    w/o SE & 77.92\% & 0.83 ms & 81.86\% & 1.48 ms \\
    per block & 78.75\% & 0.92 ms & 82.29\% & 1.58 ms  \\
    ours & 78.74\% & 0.87 ms & 82.29\% & 1.52 ms \\
\bottomrule
\end{tabular}
}
\vspace{-10pt}
\label{tab:se}
\end{table}

\paragraph{SE layer placement.} To verify the advantage of utilizing SE layers in a cross-block manner for all stages, we conduct ablation studies on ImageNet-1K by removing all SE layers (\ie, ``w/o SE'') and adopting SE layer in each block (\ie, ``per block''). As presented in \Cref{tab:se}, alternatively adopting SE layers in blocks shows a more advantageous trade-off between accuracy and latency.

\section{Conclusion}
In this paper, we revisit the efficient design of lightweight CNNs by incorporating the architectural designs of lightweight ViTs. This ends up with RepViT, a new family of lightweight CNNs for resource-constrained mobile devices. RepViT outperforms existing state-of-the-art lightweight ViTs and CNNs on various vision tasks, showing favorable performance and latency. It highlights the promising prospect of pure lightweight CNNs for mobile devices. We hope that RepViT can serve as a strong baseline and inspire further research into lightweight models.

\section{Acknowledgments}

This work was supported by National Science and Technology Major Project 2022ZD0119401, Beijing Natural Science Foundation (No. L223023), and National Natural Science Foundation of China (Nos. 62271281, 61925107, 62021002).

{
    \small
    \bibliographystyle{ieeenat_fullname}
    \bibliography{main}
}


\end{document}